\documentclass{article}

\usepackage[preprint]{neurips_2021}

\usepackage[utf8]{inputenc} 
\usepackage[T1]{fontenc}    
\usepackage{hyperref}       
\usepackage{url}            
\usepackage{booktabs}       
\usepackage{amsfonts}       
\usepackage{nicefrac}       
\usepackage{microtype}      
\usepackage{xcolor}         

\usepackage{amssymb}
\usepackage{graphicx}
\usepackage{amsmath}
\usepackage{algorithm} 
\usepackage{algpseudocode} 
\floatname{algorithm}{Pseudocode}

\algnewcommand{\LineComment}[1]{\vfill \footnotesize{\ttfamily{\textcolor{blue}{\textbackslash\textbackslash\space #1}}}}

\algrenewcomment[1]{\hfill\footnotesize{\ttfamily{\textcolor{blue}{\textbackslash\textbackslash\space #1}}}}

\usepackage{listings}

\definecolor{codegreen}{rgb}{0,0.6,0}
\definecolor{codegray}{rgb}{0.5,0.5,0.5}
\definecolor{codepurple}{rgb}{0.58,0,0.82}
\definecolor{backcolour}{rgb}{0.95,0.95,0.92}

\lstdefinestyle{mystyle}{
    backgroundcolor=\color{backcolour},   
    commentstyle=\color{codegreen},
    keywordstyle=\color{magenta},
    numberstyle=\tiny\color{codegray},
    stringstyle=\color{codepurple},
    basicstyle=\ttfamily\footnotesize,
    breakatwhitespace=false,         
    breaklines=true,                 
    captionpos=b,                    
    keepspaces=true,                 
    numbers=left,                    
    numbersep=5pt,                  
    showspaces=false,                
    showstringspaces=false,
    showtabs=false,                  
    tabsize=2
}

\hypersetup{colorlinks=true,allcolors=[rgb]{0,1,0}}

\title{Recurrent Attention Models with Object-centric Capsule Representation for Multi-object Recognition}

\author{
  Hossein Adeli$^{1*}$,\, Seoyoung Ahn$^{1}$,\, Gregory Zelinsky$^{1,2}$\\ \\\
  $^{1}$Department of Psychology, $^{2}$Department of Computer Science\\
  Stony Brook University \\
  \texttt{\{hossein.adelijelodar, seoyoung.ahn, gregory.zelinsky\}@stonybrook.edu}\\
  $^{*}$ corresponding author
  }

\begin{document}

\maketitle

\begin{abstract}
The visual system processes a scene using a sequence of selective glimpses, each driven by spatial and object-based attention. These glimpses reflect what is relevant to the ongoing task and are selected through recurrent processing and recognition of the objects in the scene. In contrast, most models treat attention selection and recognition as separate stages in a feedforward process. Here we show that using capsule networks to create an object-centric hidden representation in an encoder-decoder model with iterative glimpse attention yields effective integration of attention and recognition. We evaluate our model on three multi-object recognition tasks; highly overlapping digits, digits among distracting clutter and house numbers, and show that it learns to effectively move its glimpse window, recognize and reconstruct the objects, all with only the classification as supervision. Our work takes a step toward a general architecture for how to integrate recurrent object-centric representation into the planning of attentional glimpses. 

\end{abstract}


\section{Introduction}

Visual inputs are perceived selectively and sequentially by the brain through attention sampling processes~\citep{gottlieb2013information}. Inspired by this, a wide range of attention mechanisms have been explored and incorporated recently in Deep Learning models of vision~\citep{mnih2014recurrent, ba2014multiple, zoran2020towards,jaegle2021perceiver, xu2015show} among other domains~\citep{vaswani2017attention}. The mechanisms range from ‘soft’ highlighting of task relevant features~\citep{hu2018squeeze} 
or spatial areas~\citep{wang2017residual} to ‘hard’ glimpse-based mechanisms~\citep{elsayed2019saccader, mnih2014recurrent} inspired by fixation behavior. 
The glimpse-based mechanisms have the promise of making object recognition models more efficient and interpretable as they would only have to focus processing resources on smaller and relevant areas of the image. However, models in this domain typically treat attention selection and recognition as two separate processing stages~\citep{cordonnier2021differentiable}, first detecting the salient and relevant parts and features of the visual input before applying subsequent processing (e.g. recognition). This could limit their generalizability to domains where the relevance of certain areas and features in the image need to be determined dynamically and through ongoing object-based hypothesis formation. Understanding the dynamics of integrated attention-recognition mechanisms is important for building models that can learn to optimally sample an image, leading to more human-like and interpretable models.


Another recent development for making object recognition models more human-like and interpretable are Capsule Networks (CapsNets; \cite{sabour2017dynamic, hinton2018matrix}). CapsNets attempt to represent scenes as parse trees, with capsules in different layers representing visual entities at different levels of object granularity in the image, from small object parts in the lower levels to whole objects at the top level (with objects being represented by the corresponding category capsules). 
However, there are limitations that can prevent the wider applicability of CapsNets. First, the model does not address how visual information could be processed across multiple timesteps, as new information becomes available either through attention sampling or subsequent frames in a video. The models so far have focused on how well objects can be represented and recognized  based on a single sample of an image (but see \cite{hinton2021represent} for recent ideas on how to address this). Second, if there is one top-level capsule assigned to representing each category, how could a model represent and recognize multiple instances of the same category? 

In this work we present OCRA, an \textbf{O}bject-\textbf{C}entric \textbf{R}ecurrent \textbf{A}ttention model that combines recurrent glimpse-based attention and capsule methods. Like a CapsNet, it performs encapsulation of features to structure the higher level representations for object recognition. However, we place this structure within an encoder-decoder model with recurrent attention, thereby enabling integration of structured information across multiple attentional glimpses. Our model addresses the aforementioned limitations of the original CapsNets and shows that capsule-based binding of object features and grouping (part-whole matching) is effective in sequential detection and recognition of multiple overlapping and distinct objects. This synthesis of approaches can pave the way for building better recognition models for challenging conditions requiring a mechanism for parsing the scene to entities and a recurrent process for iterative attentional sampling and accumulation of new information.



 

\section{Related Works}
\paragraph{Soft attention models:}
Many models have been proposed that learn spatial maps or filter weights to allocate processing resources to the most relevant areas or features of an image to improve performance~\citep{chen2017sca,hu2018squeeze, park2018bam,wang2017residual}. These ``attention'' processes range from bottom-up, where each layer decides what weighting of the features to route to the next~\citep{chen2017sca, hu2018squeeze}, to top-down modulations coming from a higher level representation~\citep{cao2015look, xu2015show}. 
In this vein, transformer-based attention mechanisms~\citep{vaswani2017attention} have recently been used to create models with an integrated process of ``soft'' sampling and recognition \citep{zoran2020towards, jaegle2021perceiver}. 
Note however that soft attention models of recognition were mostly tested on images with a single prominent object, which avoids the problems of object feature binding and grouping. The sampling behavior of these models has also not yet been shown to be qualitatively similar to the fixation behavior of people, although the soft attention mechanisms could potentially capture aspects of bottom-up saliency~\citep{itti2000saliency} or feature-based attention where top-down modulations can weight the incoming representation based on task relevance~\citep{desimone1995neural, maunsell2006feature}.


\paragraph{Glimpse attention models:}
In contrast to ``soft'' attention mechanism of applying a weighting to feature maps obtained from the entire image, many other models have incorporated a process of sampling the visual input by selecting restricted ``glimpses'' of the image. Models in this domain broadly scatter along a few dimensions: whether they are differentiable~\citep{cordonnier2021differentiable,gregor2015draw} or not~\citep{mnih2014recurrent, ba2014multiple}, whether they are sequential~\citep{fu2017look} or use a single feedforward pass~\citep{cordonnier2021differentiable}, whether they are trained supervised~\citep{ba2014multiple} or self-supervised (trained to reconstruct the input objects)~\citep{eslami2016attend} or whether they are applied to images or videos (motion processing or activity recognition)~\citep{kahou2017ratm, kosiorek2017hierarchical, duta2020discovering}. DRAW~\citep{gregor2015draw} introduced a differentiable sequential spatial attention mechanism to Variational Autoencoders (VAE, \cite{kingma2013auto}) and showed that its gradual glimpse-based process improved reconstruction performance. In contrast, supervised glimpse-based models are trained directly for recognizing one or more objects in images. RAM~\citep{mnih2014recurrent} learned to move its glimpse window to sample image locations for object recognition, although its non-differentiable reinforcement learning-based attention mechanism proved difficult to train. The Saccader model applied a similar mechanism to object classification using the ImageNet dataset~\citep{elsayed2019saccader}, but in order to make the training more tractable the model preprocesses the whole image to get the class logits for all categories at all potential patch locations. While this approach can make the models more interpretable, it does not take full advantage of the glimpse behavior to efficiently sample the image as part of an integrated recognition-attention process. Glimpse selection can also be done by having a separate stage of selecting patches before feeding them to the next stage of processing (e.g. classification)~\citep{cordonnier2021differentiable}. 
These models therefore do not learn a policy for moving attention over an image that depends on the current object hypothesis, as people do in their application of object-based attention. An extension of RAM~\citep{ba2014multiple} used the whole image in a separate pathway to provide a context for glimpse selection, thereby similarly segregating the processes of attentional selection and recognition. A broad definition of glimpse models would also include models designed to have specific ``zooming in'' mechanisms for single object recognition, used to detect objects in large high-resolution images~\citep{papadopoulos2021hard} or for fine-grained classification~\citep{fu2017look}.




\paragraph{Models with recurrent and feedback connections:}
There is a growing body of work on modeling the role of feedback and recurrent connections~\citep{gilbert2013top} for different perceptual tasks~\citep{kreiman2020beyond}. Some ``task optimized'' models take an agnostic approach, adding recurrent and top-down connections and training end-to-end (similar to feedforward networks) to achieve the best performance in a task~\citep{liang2015recurrent,zamir2017feedback, kim2019disentangling}. An insight that these efforts have yielded is that the use of recurrence can allow for more compact models to have similar computational depth and reach similar levels of accuracy as bigger feedforward models~\citep{nayebi2021goal,van2020going}. Another insight from these models is that recurrence assists challenging recognition tasks, such as those with high degrees of occlusion~\citep{spoerer2017recurrent, wyatte2014early}, which is hypothesized to be due to leveraging contextualized iterative computations~\citep{van2020going}. Other work has assigned more specific roles to the recurrent and feedback connections, for example to provide hypotheses for object locations~\citep{cao2015look} or the optimal feature weightings~\citep{li2018learning} for the subsequent feedforward pass, using mechanisms that overlap with soft attention models~\citep{stollenga2014deep,wang2014attentional}. 
%

\paragraph{Object-centric models (``Slots'' and ``Capsules''):}
Objects are how people interact with the world and are therefore central to human scene understanding~\citep{scholl2001objects,spelke1990principles}. Visual objects are formed by (bottom-up) part-whole matching and Gestalt processes interacting with (top-down) objectness priors and knowledge of object categories~\citep{greff2020binding, vecera2000toward, wagemans2012century}. Object-centric models use these processes to discover objects and segregate their representations into different ``slots''~\citep{greff2020binding, goyal2019recurrent}. Attention mechanisms have played a major role in object-centric models by enabling the iterative discovery and representation of an object’s properties. 
AIR~\citep{eslami2016attend} used a spatial transformer component~\citep{jaderberg2015spatial} to attend to objects, infer their properties and reconstruct them sequentially. 
MONET~\citep{burgess2019monet} first creates spatial object masks that are then used sequentially to reconstruct each entity using VAEs, effectively separating the attention routing and representation learning processes into two stages. This approach has been shown to be effective for downstream object reasoning tasks~\citep{ding2020object}. 
Although not using sequential spatial attention, some other models in this domain relatedly employ iterative processes to dynamically route and segregate objects to different slots~\citep{greff2019multi, locatello2020object}.
Capsule networks~\citep{hinton2018matrix, sabour2017dynamic} are slot-based models that have capsule slots for each category placed at the top of a hierarchy of capsules representing a scene parse tree. 
Similar to other object-centric models, the entity encapsulation performed by CapsNets, creates object-centric representations that can lead to better downstream task performance~\citep{qin2020deflecting}.


\section{Approach}


Our approach is object-centric through the use of the entity encapsulation property of CapsNets, using it to structure representations for classification and attention planning. It is also consistent with soft attention and recurrent models of recognition. Adding glimpses to these models of recognition enables our model not only to leverage recurrent computation, but also to select the most relevant incoming visual inputs. OCRA therefore intersects all of the discussed modeling approaches, which we demonstrate to be key to its performance.  


OCRA's attention mechanism builds on the DRAW architecture~\citep{gregor2015draw}, which we found to be cognitively plausible. The ``attention window'' is a grid of filters applied to a small or a large area of the image. However, because the number of filters covering the attention window remains constant, as the window gets bigger it samples increasingly low-resolution information, creating a tradeoff between the size of the attention window and the resolution of information extracted. This property is aligned with ``zoom lens'' theories of human attention~\citep{eriksen1986visual, muller2003physiological}, that similarly propose a variable-sized attention process that can be broadly or narrowly allocated to an input, but one that is constrained by the same trade off.
The original DRAW model~\citep{gregor2015draw} was formulated as an autoregressive VAE, as it was trained for stepwise self-supervised reconstruction. 
In contrast, our formulation uses a deterministic encoder-decoder approach to make the model easier to train using both classification and reconstruction losses, because we want it to predict both the category classification based on latent class capsules, and image reconstruction based on decoder output. 


\subsection{OCRA Architecture}
\vspace*{-2mm}

\begin{figure}[!h]
    \centering
    \includegraphics[width=\textwidth]{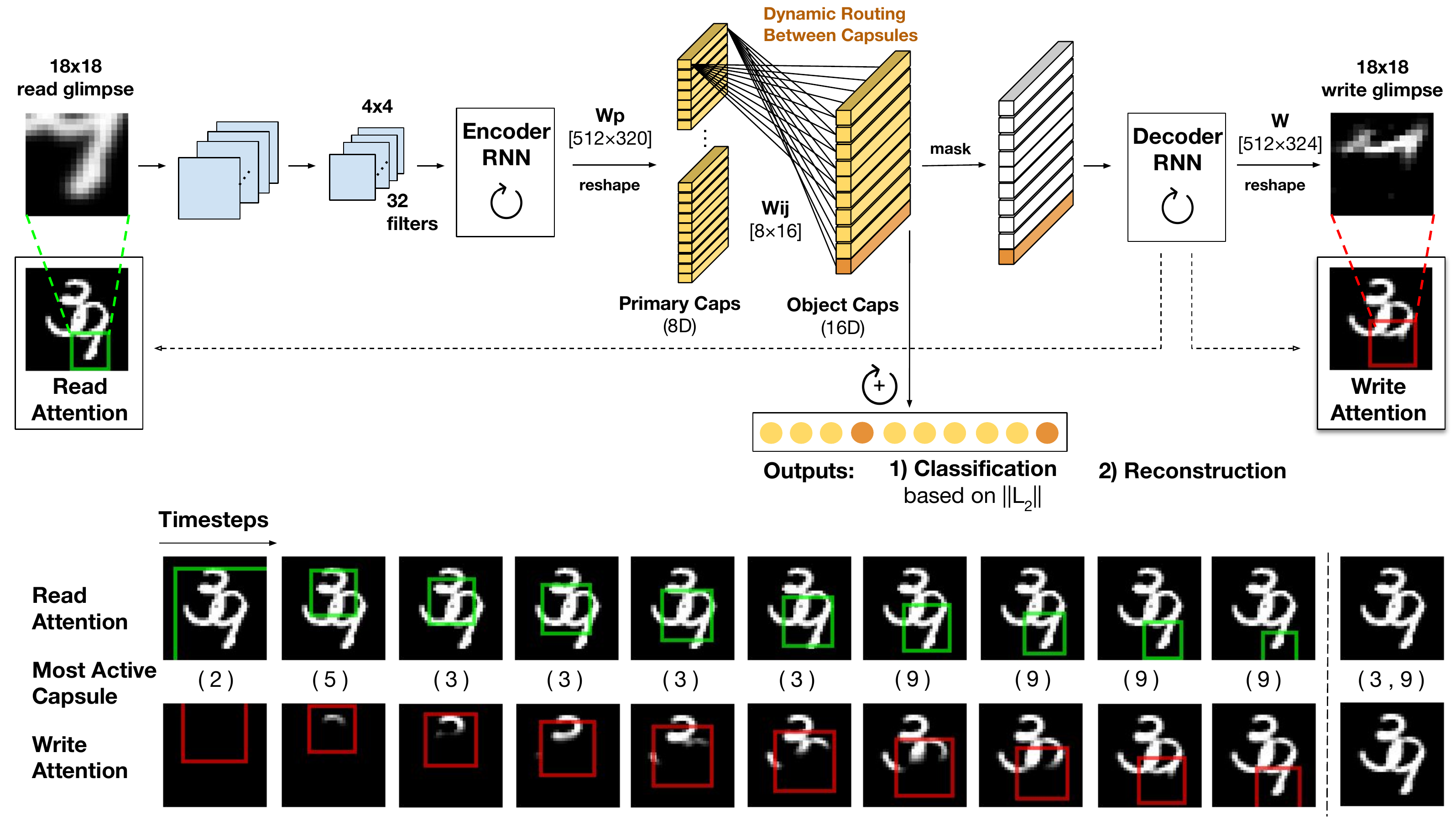}
    \caption{OCRA performing multi-object recognition on an overlapping-digit image.}
    \label{fig:modelpipeline}
    
\end{figure}

The OCRA architecture is shown in Fig.~\ref{fig:modelpipeline}. At each timestep, a new glimpse from the image is encoded by a hierarchy of modules to yield a structured latent representation. The decoder then reconstructs the glimpse using this latent representation. The encoder-decoder steps are taken sequentially, determining the attention glimpse location for the following step. 
We provide an overview of the OCRA components and loss functions below. Pseudocodes and implementation details are provided in the Supplementary~\ref{sec:implementation}. A pytorch~\citep{paszke2019pytorch} implementation of OCRA with additional details and results are provided in this repository: \href{https://github.com/Hosseinadeli/OCRA}{github.com/hosseinadeli/OCRA}. 

\paragraph{Read and Write Attention:}
\label{paragraph:attention}

At each timestep a glimpse, $g_t$, is sampled through applying a grid of N $\times$ N Gaussian filters on the input image $x$. We set the glimpse size to 18$\times$18 for our experiments, with a sample glimpse shown in Fig.~\ref{fig:modelpipeline} left. The Gaussian filters are generated using four parameters: $g_X, g_Y, \delta, \sigma^2$, which specify the center coordinates of the attention window, the distance between equally spaced Gaussian filters in the grid, and the variance of the filters, respectively. 
All of these parameters are computed via a linear transformation of the previous step decoder RNN (Recurrent Neural Network) output $h^{dec}_{t-1}$ using a weight matrix $W_{read}$, which makes the attention mechanism fully-differentiable. A similar procedure applies to the \textit{write} attention operation. The decoder RNN output $h^{dec}_t$ is linearly transformed into an M $\times$ M write patch $w_t$ (set to 18 $\times$ 18 in our experiments), which is then 
multiplied by the Gaussian filters to reconstruct the written parts in the original image size (Fig.~\ref{fig:modelpipeline} right). 
The Gaussian filters used for the \textit{write} operation differed from those used for the \textit{read} operation, and were computed from four parameters obtained from a separate linear transformation, $W_{write\_attention}$, of the decoder RNN output $h^{dec}_t$.
Detailed algorithms for Read and Write attention operations are provided in Pseudocode~\ref{alg:read} and \ref{alg:write} in the Supplementary along with an illustration of the attention mechanisms (\ref{sec:glimpse_attn}).



\paragraph{Encoder:}
\label{encoder}
After a glimpse is selected from the input image by the read attention operation, it is processed first using a two-layer convolutional neural network (CNN) with 32 filters in each layer. Kernel sizes were set to 5 and 3 respectively for the first and the second layers. Each convolutional layer is followed by max pooling with a kernel size of 2 and a stride of 2, and Rectified Linear Units (ReLU) were used for non-linear activation functions. Given the glimpse size of 18 $\times$ 18, the resulting 32 feature maps are of size 4 $\times$ 4. The feature maps, $g^{conv}_t$, are reshaped (to a vector of size 512) and used as input to the encoder Recurrent Neural Network (RNN), along with the encoder RNN hidden state from the previous step; $h^{enc}_{t-1}$. We used LSTM~\citep{hochreiter1997long} units (size 512) for the recurrent layers in our model.  




\paragraph{Latent Capsule Representations and Dynamic Routing:}
\label{sec:capsule}
We use a vector implementation of capsules~\citep{sabour2017dynamic} where the length of the vector represents the existence of the visual entity and the orientation characterizes its visual properties. The primary level capsules are generated through a linear read out of the encoder RNN; $h^{enc}_t$. These capsules are meant to represent lower-level visual entities (``parts'') that belong to one of the higher-level capsules in the object capsule layer (``whole"). To find this part-whole relationship, we used the dynamic routing algorithm proposed by \cite{sabour2017dynamic}. Dynamic routing is an iterative process where the assignments of parts to whole objects (coupling coefficients) are progressively determined by agreement between the two capsules (measured by the dot product between the two vector representations). 
For example, if the prediction for a digit capsule $j$ from a primary capsule $i$, ($\hat{p}^{j|i}_t \leftarrow W^{ij}_t p^{i}_t$), highly agrees with the computed digit capsule $(\sum_{i}c^{ij}_t\hat{p}^{j|i}_t)$, the coupling coefficient $c^{ij}_t$ is enhanced so that more information is routed from primary capsule $i$ to object capsule $j$. 
Coupling coefficients are normalized across the class capsule dimension following the max-min normalization~\citep{zhao2019capsule} as in the Supplementary Eq.~\ref{eq:maxmin}. 
This routing procedure iterates three times. We used this method instead of the softmax normalization in~\cite{sabour2017dynamic} because we observed the latter method would not differentiate between the coupling coefficients. In our experiments we used 40 primary level capsules, each a vector of size 8. The object capsules are vectors of size 16 and there are 10 of them corresponding to the 10 digit categories. For the object level capsules, we use a squash function (the Supplementary Eq.~\ref{eq:squash}) to ensure that its vector length is within the range of 0 to 1 to represent the probability of a digit being present in the glimpse at each step. Once the routing is completed, we compute the vector length (L2 norm) of each object capsule to obtain classification scores. The final digit classification is predicted based on the scores accumulated over all timesteps. Algorithms are provided in Pseudocode~\ref{alg:capsule}.




\paragraph{Decoder:}
\label{decoder}
The object capsules provide a structured representation that can be used for decoding and glimpse selection. We first mask the object capsules so that only the vector representation from the most active capsule is forwarded to the decoder RNN, which also inputs the hidden state from the previous step, $h^{dec}_{t-1}$.  Because the decoder maintains through recurrence the ongoing and evolving object-based representation of the image, it is best suited to determine the next read glimpse location (as discussed earlier). The state of the decoder RNN is also used through two linear operations to determine what and where to write in the canvas to reconstruct the image. 


\subsection{Loss Function}
OCRA outputs object classification scores (cumulative capsule lengths) and image reconstruction (cumulative write canvas). Losses are computed for each output and combined with a weighting as in Eq.~\ref{eq:totalloss}.
For reconstruction loss, we simply computed the mean squared differences in pixel intensities between the input image and the model's reconstruction. For classification, we used margin loss (Eq.~\ref{eq:marginloss}). For each class capsule j, the first term is only active if the target object is present ($T_j > 0$) where minimizing the loss pushes the capsule length to be bigger than target capsule length minus a small margin (m). The second term is only active when the target capsule length is zero and in that case minimizing the loss pushes the predicted capsule length to be below a small margin (m). For all the experiments in this paper we used Adam optimizer~\citep{kingma2014adam}. See Supplementary~\ref{sec:loss} for a detailed explanation and the pseudocode for the loss functions. 

\begin{gather}
    \label{eq:totalloss}
    Total \, Loss =  \sum_{j \in class} Class\,Loss_j + \lambda_{recon} Recon\, Loss \\
    \label{eq:marginloss} 
    \begin{aligned}
    Class \,Loss = \sum_{j \in class} max(0,min(T_j,1)) \cdot max(0, (T_j-m)-\|d_j\|)^2 + \\ \lambda_{absent}\cdot max(0, 1-T_j)\cdot max(0, \|d_j\|-m)^2 
    \end{aligned}
\end{gather}



\section{Results}

Attention is hypothesized to be helpful in difficult recognition tasks, involving multiple (small) objects or difficult feature discrimination, heavily occluded objects, or objects appearing against noisy object-similar backgrounds. Considering this, here we use three proof of concept multi-object recognition tasks to illustrate the effectiveness of our proposed attention-recognition mechanism. The MultiMNIST task~\citep{sabour2017dynamic} tests the model's recognition performance on highly overlapping digits. We hypothesize that our object-based attention sampling, paired with recurrent processing, will be effective in recognizing objects despite high degrees of occlusion. The second task uses the MultiMNIST cluttered dataset~\citep{ba2014multiple}. This task tests the model's ability to learn to make attention glimpses to individual distant objects while ignoring object-similar background clutter. We selected these two MultiMNIST tasks to make sure that individual objects can be recognized easily and that the difficulty of these (still challenging) tasks would be in having multiple objects, occlusions and distractors, to be able to isolate the effectiveness of the proposed object-centric recurrent attention mechanism. Also these two tasks require different attention mechanisms and to our knowledge no model has been shown to perform both at a high level. The third task is sequence prediction on the Street View House Numbers dataset (SVHN)~\citep{goodfellow2013multi} and can test our model on its ability to scale up to more real-world datasets with a different number of objects in images. 

All model accuracies presented in this section are averages of 5 runs (see~\ref{sec:training_supp} for details of training and hyperparameters selection for different experiments). All error rates are measured on image level, meaning that the response is incorrect if any object in the image is not recognized correctly. 
\subsection{MultiMNIST}


\begin{figure}[b]
    \centering
    \vspace*{-1mm}
    \includegraphics[width=\textwidth]{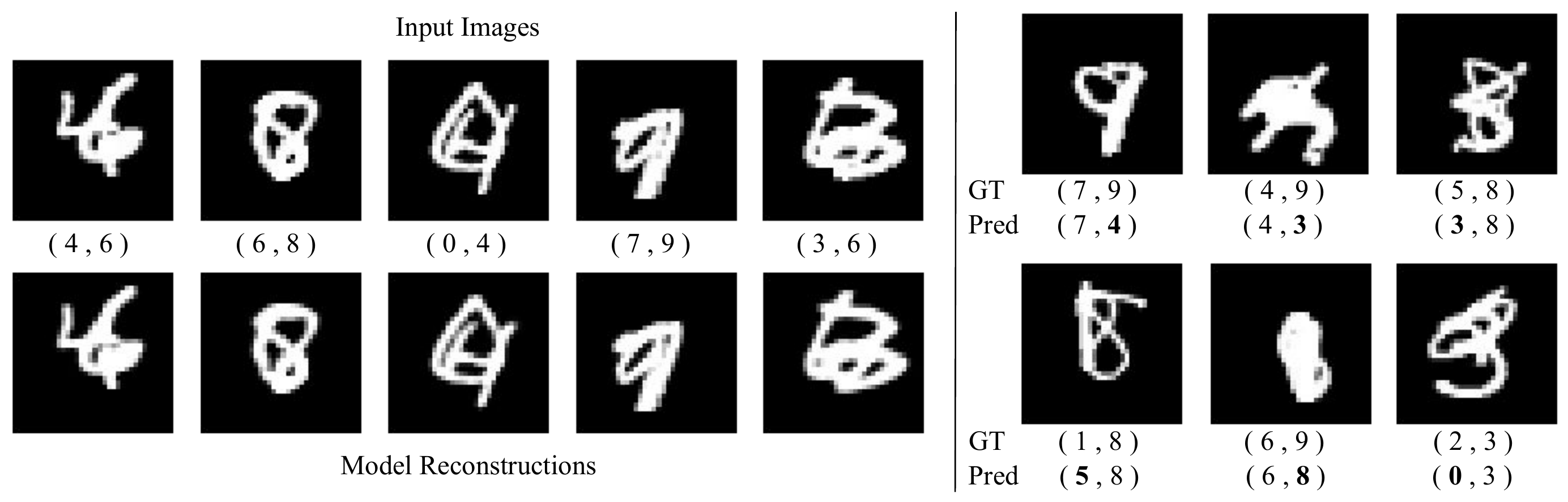}
    \vspace*{-1mm}
    \caption{Examples of OCRA outputs on MultiMNIST classification (error cases on the right)} 
    \label{fig:OCRA_mm_samples}
    \vspace*{-1mm}
\end{figure}

\begin{table}[t]
\caption{MultiMNIST classification error rates. The error rates for the CapsNet and the CNN baseline are taken directly from Sabour et al., (2018)}
\label{tab:multiMNIST}
    \centering
    \begin{tabular}{lccc}
    \specialrule{1.2pt}{1.5pt}{1.5pt}
    Model & Model size & Train size & Error rate \\ \specialrule{1.2pt}{1.5pt}{1.5pt}
    CNN Baseline & 24.56M & 60M & 8.1\% \\ \midrule
    CapsNet (Sabour et al., 2018)  & 11.36M & 60M & 5.2\%  \\ \specialrule{1.2pt}{1.5pt}{1.5pt}
    OCRA-3glimpse  & 3.87M & 3M & 7.24\% ($\pm$0.11)  \\ \midrule
    OCRA-10glimpse & 3.87M & 3M & 5.08\% ($\pm$0.17)  \\ 
    \specialrule{1.2pt}{1.5pt}{1.5pt}
    \end{tabular}
    \vspace*{-4mm}
    \centering
\end{table}

\paragraph{MultiMNIST Dataset:}
We generate the MultiMNIST dataset following the method from \cite{sabour2017dynamic}. Each image in this dataset contains two overlapping digits sampled from different classes of the MNIST hand-written digits dataset~\citep{lecun1998gradient} (size 28$\times$28 pixels). After the two digits are overlaid, each is shifted randomly up to 4 pixels in horizontal and vertical directions, resulting in images of size 36$\times$36 pixels with on average 80\% overlap between the digit bounding boxes. We generated 3M images for training, and 500K images for testing, and ensured that the training/testing sets were generated from the corresponding MNIST sets (i.e., the training set in MultiMNIST was only generated from the training set in MNIST). See Fig.~\ref{fig:OCRA_mm_samples} for examples.


\subsubsection{MultiMNIST Results}

Fig.~\ref{fig:OCRA_mm_samples} (left) shows some samples of the model making correct predictions and the resulting reconstructions. The 6 images on the right show all the errors that OCRA made in one test batch (size 128). Table~\ref{tab:multiMNIST} shows the OCRA accuracy for this task compared to two competing models. Our model with 10 timesteps outperforms the CapsNet model \cite{sabour2017dynamic} while having only a third of the number of parameters. Moreover, the training set for our model is 20 fold smaller (3M compared to 60M), and in contrast to the CapsNet model ours does not use individual segmentations for each object during training. We saw a clear effect of the number of timesteps in our model, with the error rate dropping from 3 timesteps to 10 timesteps. Fig.~\ref{fig:modelpipeline} (bottom) shows the glimpse behavior on a sample image. The model starts with a more global glimpse but then moves its attention window, first to one object and then to the other, recognizing and reconstructing each sequentially. The gradual spreading of the reconstruction shown (bottom row) is consistent with the spreading of attention within an object hypothesized by object-based models of attention~\citep{jeurissen2016serial, ekman2020object}. The most active capsule at each timestep is indicated by the digits in between the two rows. 


\subsubsection{Ablation experiments}

In this section we start from the OCRA-3glimpse model and probe different model components; the object-centric representation, the glimpse mechanism and the recurrent processing; to measure their impact on accuracy. Results are shown in Table~\ref{tab:multiMNIST_ablation}.

\paragraph{The effect of recurrent attention mechanisms:}

In the first ablation experiment we asked how well 
a recurrent model using object-centric representation would perform without the ability to sample glimpses. OCRA-Recurrent (Table~\ref{tab:multiMNIST_ablation}) performs multi-step processing on the input image using the recurrence in its encoder and decoder RNNs but lacks the ability to glimpse at specific locations. Therefore the model receives the entire image as input at each processing step for which it requires a bigger number of parameters (36$\times$36 pixel input in contrast to 18$\times$18 pixel glimpse). We then trained this model using three timesteps to make it comparable to OCRA-3glimpse. As shown in Table~\ref{tab:multiMNIST_ablation}, this model performs worse than OCRA-3glimpse highlighting the important role of glimpse mechanism in our model performance. We then asked what if we removed the recurrent attention mechanism completely from our model. What is left is a model that makes one feedforward pass with the full resolution image as its input, binds features for either objects in separate category capsules, and then feeds them to the decoder (without masking the object capsules) to reconstruct the whole image at once. This is the Feedfoward model in Table~\ref{tab:multiMNIST_ablation}. Similar to the OCRA-recurrent, having the model input the whole image in full resolution requires a bigger backbone and a larger number of parameters. This model shows higher error rate compared to the OCRA-recurrent supporting previous work on how recurrent dynamics can assist recognition tasks high degrees of occlusion~\citep{spoerer2017recurrent, wyatte2014early}.
Taken together, the results show that while recurrent computation can be effective for challenging recognition tasks with high degrees of occlusion, when it is paired with an attention glimpse mechanism more compact and better performing models can be built. 

\begin{table}[t]
\caption{Ablation study results on MultiMNIST classification}
\label{tab:multiMNIST_ablation}
    \centering
    \begin{tabular}{lcc}
    \specialrule{1.2pt}{1.5pt}{1.5pt}
    Model & Model size  &  Error rate \\ \specialrule{1.2pt}{1.5pt}{1.5pt}
    OCRA-3glimpse & (3.87M) & 7.24\% ($\pm$0.11) \\ \specialrule{1.2pt}{1.5pt}{1.5pt}
    Bigger models with the whole image as input \\(no glimpse mechanism) 
    \vspace*{1mm} \\
    OCRA-Recurrent-3step &  (8.58M) & 8.98\% ($\pm$0.14)  \\ \midrule
    -Feedforward & (6.47M) & 10.63\% ($\pm$0.10) \\ 
    \specialrule{1.2pt}{1.5pt}{1.5pt}
    OCRA-3glimpse with 1 routing step & (3.87M) & 7.77\% ($\pm$0.24) \\ \midrule
    -3glimpse without capsule representation & (3.87M) & 8.04\% ($\pm$0.13) \\ \specialrule{1.2pt}{1.5pt}{1.5pt}
    \end{tabular}
    \vspace*{-2mm}
    \centering
\end{table}

\paragraph{The role of capsule architecture and dynamic routing between capsules:}
We performed two ablation studies to examine the impact of capsule architecture on model performance. 
We first asked whether the multiple dynamic routing steps between the two capsule layers has an impact on the error rate since the efficiency of the routing mechanism used by CapsNets has had mixed results to date~\citep{gu2021capsule, tsai2020capsules, wang2018optimization, zhao2019capsule}. As shown in Table~\ref{tab:multiMNIST_ablation}, decreasing the routing step to 1 (uniform coupling coefficients) resulted in increased error rate compared to OCRA-3glimpse which had three dynamic routing steps. We then examined the role of Capsules in the model accuracy, by replacing that architecture with two fully connect layers (with the same size as two capsule layers; 320 and 160 units) and a classification readout. The new model still iteratively processes the image and the classification scores are read out from the second fully connected layer representation at each time step, which are then all combined across timesteps to make the final classification decision.  As shown in Table~\ref{tab:multiMNIST_ablation}, we saw further increase in the error rate compared to the previous model showing the effect of encapsulation of information for performing this task. While these effects offer evidence for the role of capsule architecture and dynamic routing, they are relatively smaller compared to the ones from ablating the the recurrent attention. This suggests that the models can to some extent compensate for smaller routing steps or removal of the capsule architecture on this task. This could be due to two factors. First, the dynamic routing mechanism can flexibly learn to route information in a single routing step, allowing the weights between the capsule layers to do the binding without the need for multiple refining steps. Second, because the model has a glimpse mechanism it can route information globally, thereby potentially reducing the benefit derived from feature encapsulation and local dynamic routing. 

\subsection{MultiMNIST Cluttered}
In this section we show that the attention mechanism in OCRA can handle noisy backgrounds and that its capsules can be used to recognize multiple objects from the same category. 

\paragraph{MultiMNIST Cluttered dataset:} We generated the MultiMNIST Cluttered dataset, similar to the cluttered translated MNIST dataset from~\cite{mnih2014recurrent}. For each image, 2 digits and 6 digit-like clutter pieces are placed in random locations on a 100$\times$100 blank canvas. All digits were sampled from the original MNIST dataset~\citep{lecun1998gradient} and the two digits in each image could be from the same or different categories. Clutter pieces were generated from other MNIST images by randomly cropping 8$\times$8 patches. We generated 180K images for training and 30K for testing, ensuring to maintain the same MNIST training/testing separation. 

\begin{figure}[!h]
    \centering
    \includegraphics[width=\textwidth]{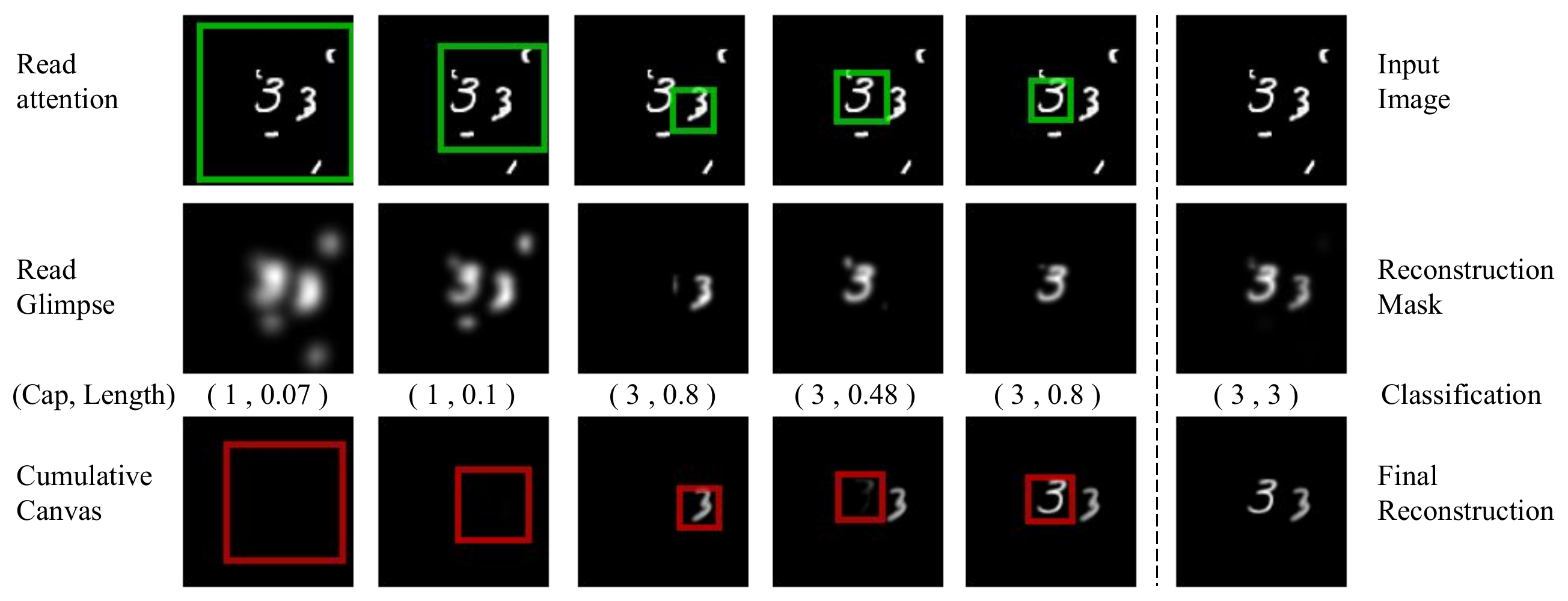}
    \vspace*{-4mm}
    \caption{Attention, recognition and reconstruction processes of OCRA on MultiMNIST Cluttered}
    \label{fig:OCRA_mmc_process}
\end{figure}

\subsection{MultiMNIST Cluttered results}

For this task we added a background capsule to the 10 class capsules, similar in approach to~\cite{qin2020deflecting}). This gives the model the choice to dynamically route background noise in its attention windows to a non-class capsule, thereby allowing the model to exclude the noise from its object representations. 
We also define a reconstruction mask that is the averaged sum of all the read glimpses converted into image dimensions (Fig.~\ref{fig:OCRA_mmc_process} middle row, right). We integrate this mask into our loss function by multiplying it by the input image before comparing it to the model reconstruction. This mask effectively focuses the loss so that the model is accountable for reconstructing only the areas where it had glimpsed, allowing the model to be selective with its glimpses and write operations. 

\begin{figure}[b]
    \centering
    \vspace*{-1mm}
    \includegraphics[width=\textwidth]{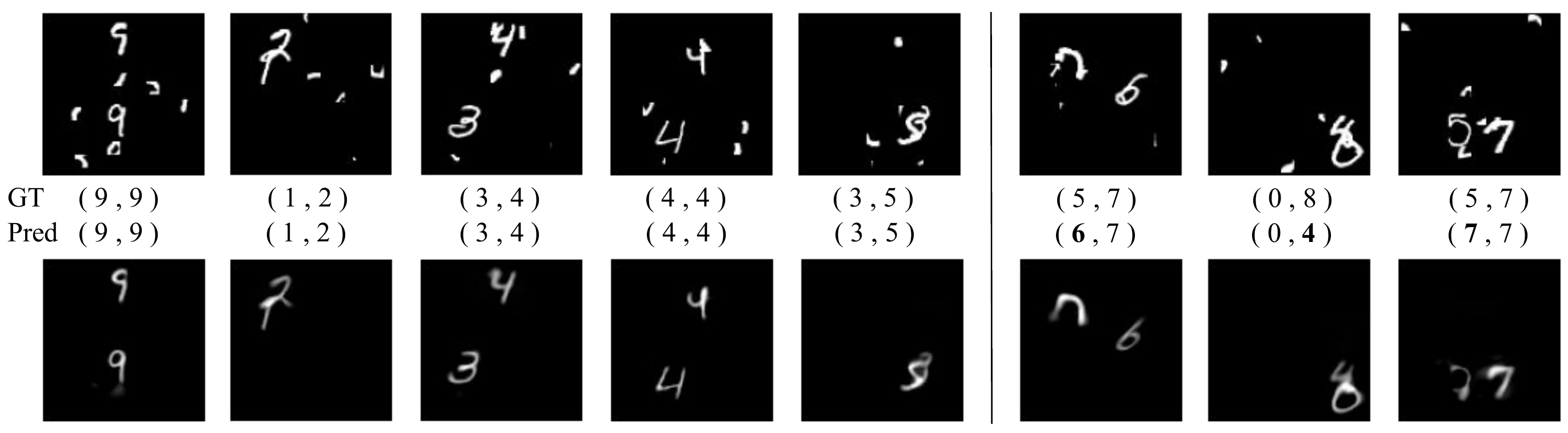}
    \caption{OCRA MultiMNIST cluttered sample predictions for recognition and reconstruction (errors are in bold on the right side)}
    \label{fig:OCRA_mmc_samples}
\end{figure}

Fig.~\ref{fig:OCRA_mmc_process} shows OCRA performing detection, classification and reconstruction of the digits in five timesteps for a sample image (top right) from this task. The top row shows the attention windows and the middle row shows the read glimpse at each step. The glimpses are converted to image dimensions for creating the reconstruction mask, as seen on the middle right. The most active capsule that is routed to the decoder at each step, and its length, are provided above the cumulative canvas (bottom row). Utilizing the reconstruction mask, the model, with only classification supervision, learns that the best strategy is to move its glimpse to digits and to write to the canvas only when it is confident of the digit classification. The model leverages the object-centric representation for attention planning and reconstruction, reflected in the behavior of clearly selecting objects and ignoring the distractors.  We provide many more examples of the model behavior in the github repo. Fig.~\ref{fig:OCRA_mmc_samples} shows more examples of model predictions for both recognition and reconstruction, with correct responses on the left side and errors on the right. Most errors by OCRA on this task are due to a digit overlapping with the other digit or noise pieces in ways that change their appearance from the underlying ground truth.




\begin{table}[t]
\caption{MultiMNIST Cluttered classification error rates. The error rates for other models are taken directly from the corresponding papers (no error bars were provided)}
\label{tab:MultiMNIST_cluttered}
    \centering
    \begin{tabular}{lc}
    \specialrule{1.2pt}{1.5pt}{1.5pt}
    Model & Error rate \\ 
    \specialrule{1.2pt}{1.5pt}{1.5pt}
    RAM & 9\% \\ \midrule
    DRAM w/o context & 7\% \\\midrule
    DRAM & 5\% \\
    \specialrule{1.2pt}{1.5pt}{1.5pt}
    OCRA-5glimpse & 8.37\% ($\pm$1.43) \\ \midrule
    OCRA-7glimpse & 7.12\% ($\pm$1.05) \\ 
    \specialrule{1.2pt}{1.5pt}{1.5pt}
    \end{tabular}
\end{table}

Table~\ref{tab:MultiMNIST_cluttered} shows the classification results on this task. Two versions of the DRAM model~\citep{ba2014multiple} are included, with the one utilizing a context network performing best among the baseline models with a 5 percent error rate (taken directly from the reference paper where no error bars were provided). However, the context network in this model inputs the whole image in a separate pathway to plan attention selection, thereby separating it from the recognition pathway. OCRA has one pathway and strings together glimpses from multiple steps to have an integrated recognition and attention planning mechanism. Our model does this through its ``zoom lens'' attention processes that can switch between local and global processing as needed. The early glimpses in the model taken from the whole image, even though they are low resolution gists (Fig.~\ref{fig:OCRA_mmc_process}, middle row, left), allow the model to plan its future glimpses. Later glimpses are focused on individual objects to mediate the recognition and reconstruction processes. OCRA's performance on this task is comparable to the baseline models, despite the model being much smaller in size (DRAM has over 10M parameters) and easier to train (given its differentiable attention mechanism). We performed a similar ablation study to the previous task here by replacing the CapsNet architecture with two fully connected layers and a classification read out. We were not able to achieve error rates below 20 percent for the resulting model, indicating how essential the object-centric representation of CapsNet is for successfully performing this task.

\subsection{SVHN}

\begin{figure}[!h]
    \centering
    \vspace*{-1mm}
    \includegraphics[width=\textwidth]{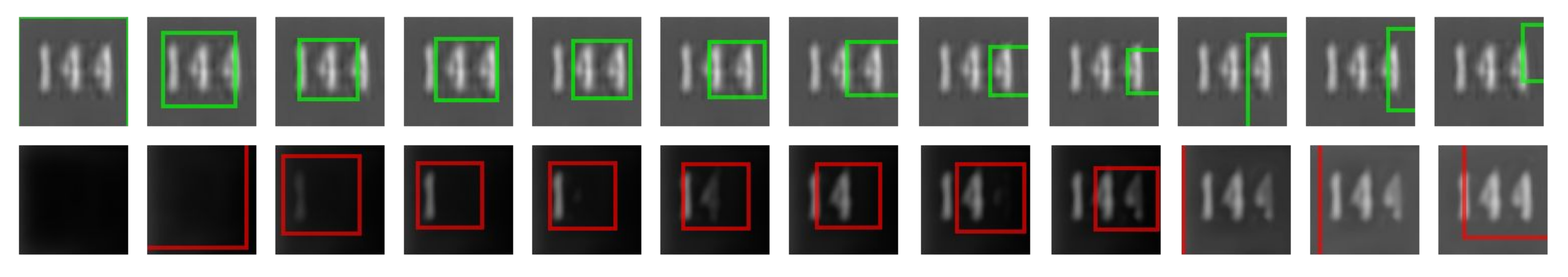}
    \vspace*{-5mm}
    \caption{OCRA glimpse behavior (top row) and cumulative canvas (bottom) on an SVHN image}
    \label{fig:OCRA_multisvhn}
    \vspace*{-1mm}
\end{figure}

Street View House Numbers (SVHN) dataset~\citep{goodfellow2013multi} consists of  real-world images of house numbers, each containing a sequence of one to five digits. This dataset tests OCRA on both its applicability to more complex real-world stimuli and also on handling a varying number of objects in an image with multiple instances of any category. The original dataset is divided to train, test and extra images. We combined the train and extra sets to create a bigger training set and also converted the images to grayscale following~\citep{ba2014multiple}, resulting in a train set of size 235K (10 percent for validatoin) and test set of size 13K. We made two changes to the model for this task. First, we increased the number of convolutional filters in the backbone from 32 to 64 in each of the two layers. Second, we added a readout layer to predict the digits in a sequence based on the capsule lengths as the model makes its pass across the image. The resulting model had 5.1 Million parameters. We train the model to "read" the digits from left to right by having the order of the predicted sequence match the ground truth from left to right. We allow the model to make 12 glimpses, with the first two not used for readout and the object capsule lengths from every following two glimpses will be read out for the output digits (e.g. the capsule lengths from the 3rd and 4th glimpses are read out to predict digit number 1; the left-most digit; and so on). Fig.~\ref{fig:OCRA_multisvhn} shows the model behavior for a sample image, as it glimpses, recognizes and reconstructs the objects in the sequence. The model achieved 2.65\% ($\pm$0.11) error rate on recognizing individual digits and 10.07\% ($\pm$0.53) error rate for recognizing whole sequences. We believe exploring bigger convolutional backbones (compared to a simple two-layer CNN we used) would be essential for improving the sequence prediction accuracy. More examples of model behavior and some classification errors are provided in the github repo. 

\section{Discussion}
\label{sec:discussion}

We believe visual perception to be a sequential process, but one that requires the integration of attention and recognition. While models might use pre-processing or patch-selection mechanisms to solve specific tasks, methods should strive to capture a more dynamic integration of attention and recognition to be able to replicate human-level performance. We also believe that this modeling approach can be used in cognitive and neuroscience research to inform debates on the role of different connections in the visual system (feedforward, recurrent and feedback) for different tasks.  



OCRA builds on capsule methods~\citep{sabour2017dynamic}, recurrent attention and DRAW architecture \citep{gregor2015draw}. While it addressed some of their limitations, there is still room for improvement. For one, we believe that a better capsule routing algorithm will be important in applying our method to more complex objects that require more complex part-whole matching. Also using a bigger backbone and generally a bigger architecture will be crucial. It is a direction for future work to determine whether our approach will scale up well to more complex objects and scenes. 

We made architecture choices in OCRA to generally align with constraints known to exist in visual perception.  We believe that these constraints, if captured correctly, would not only have the potential to improve model performance, but will also allow the model to predict human fixation behavior. We further see a broader application of our model in understating and modeling expert attention. For example, in the context of medical imaging and cancer screening by radiologists~\citep{shen2021interpretable, mall2018modeling, pesce2019learning}, a model of experts' attention and recognition behavior could lead to a greater understanding of underlying reasons for errors (misses and false positives). Moreover, having more interpretable models that could be build using effective glimpse attention and part-whole matching mechanisms can mitigate the risk of these expert models in practice.

\bibliography{neurips_2021}


\clearpage
\appendix
{\huge\textbf{Supplementary Material}}

\section{Method Details}
\label{sec:implementation}

\subsection{Pseudocode for OCRA Overview}

\begin{algorithm}[H]
	\caption{OCRA Architecture Overview} 
	\label{alg:ocra}
	\begin{algorithmic}[1]
	    \State Initialize encoder and decoder RNN hidden states, $h_{0}^{enc}, h_{0}^{dec}$, to zero
	    \State Initialize classification score vector $s_0$ and reconstruction $canvas_0$ to zero
		\For{$t$ in $timesteps=1,2,\ldots$}
		    \State Get read glimpse: $g_t \leftarrow READ(x, h^{dec}_{t-1})$
		    \Comment {read operation in Pseudocode~\ref{alg:read}}
		    
		    \State Apply two convolutional layers with max pooling: $g^{conv}_t \leftarrow\,CNN(g_t)$
		    
		    \State Update encoder RNN's hidden representations: $h^{enc}_t \leftarrow RNN^{enc}(h^{enc}_{t-1}, g^{conv}_t)$
		    
		    \State Compute object capsules through dynamic routing: $d_t \leftarrow CAPSULE(h^{enc}_t)$
		    \Comment{capsule operation in Pseudocode~\ref{alg:capsule}}
		    \State Accumulate classification score for digit $j$: $s_t^j \leftarrow s_{t-1}^j + \| d^j_t \|$ 
    		\State Mask with zeros all but the digit capsule with max classification score: $d^{mask}_t \leftarrow mask(d_t)$
    		
		    \State Update decoder RNN's hidden representations: $h^{dec}_t \leftarrow RNN^{dec}(h^{dec}_{t-1}, d^{mask}_t)$
		    \State Reconstruct image with write attention:  $w_t \leftarrow WRITE(h^{dec}_t)$
		    \Comment {write operation in Pseudocode~\ref{alg:write}}
		    
		    \State Update reconstruction canvas: $canvas_t = canvas_{t-1} + w_t$
		\EndFor
	\end{algorithmic} 
\end{algorithm}

\subsection{Pseudocode for Capsule Representation and Dynamic Routing}
The dynamic routing algorithm performs as follows: At each routing step, each primary level capsule $i$ provides a prediction for each object level capsule $j$. These predictions are then combined using the coupling coefficients $c^{ij}$ to compute the object level capsules. Then the agreement (dot product) between the object level capsules and the predictions from each primary level capsule impacts the coupling coefficients for the next routing step.
\begin{algorithm}[H]
	\caption{Capsule Representation and Dynamic Routing} 
	\label{alg:capsule}
	\begin{algorithmic}[1]
	    \State $h^{enc}_{t}$ = current encoder representation 
	    \Procedure{Capsule}{$h^{enc}_t$}
		    \State Apply linear transform on $h^{enc}_t$ to create primary capsule: $p_t \leftarrow {W_p}(h^{enc}_t)$
		    \State For all object capsule $j$ and primary capsule $i$ initialize $b^{ij}_t$ to zero
		    \For{$r$ in $routings=1,2,\ldots$}
		        \State For all object capsule $j$ compute prediction from all primary capsule $i$: $\hat{p}^{j|i}_t \leftarrow W^{ij}_t p^{i}_t$
		        \State For all object capsule $j$: $d^j_t \leftarrow squash(\sum_{i}c^{ij}_t\hat{p}^{j|i}_t)$
		        \Comment{squash function in Eq.~\ref{eq:squash}}
		        \State For all object capsule $j$ and primary capsule $i$: $b^{ij}_t \leftarrow b^{ij}_t + \hat{p}^{j|i}_t\cdot{d^j_t}$  
		        \State Max-Min Normalize: $c^{ij}_t \leftarrow maxmin(b^{ij}_t)$
		        \Comment{maxmin function in Eq.\ref{eq:maxmin}}
		    \EndFor
		    \State For all object capsule $j$: $d^j_t \leftarrow squash(\sum_{i}c^{ij}_t\hat{p}^{j|i}_t)$
		\EndProcedure
	\end{algorithmic} 
\end{algorithm}

\subsection{Max-Min Normalization}
In the original CapsNets~\citep{sabour2017dynamic}, the softmax normalization was used for normalizing coupling coefficients in dynamic routing. However, the softmax operation fails to differentiate between the coupling coefficients. To remedy this, we used max-min normalization that is proposed in~\cite{zhao2019capsule}, applying it after each routing operation. Lower- and upper-bounds for normalization, $lb$ and $ub$, were set to 0.01 and 1.0, respectively. 


\begin{equation}
    \label{eq:maxmin}
     c^{ij}_t = lb + (ub-lb)\frac{c^{ij}_t-min(c^{ij}_t)}{max(c^{ij}_t)-min(c^{ij}_t)}
\end{equation}

\subsection{Squash Function}
We use a vector implementation of capsules~\citep{sabour2017dynamic} where the length of the vector represents the existence of the visual entity and the orientation characterizes its visual properties. To ensure that the capsule vector length ranges from 0 to 1, e.g., 0 for absence and 1 for presence, we used the non-linear squash function as below. The $v^j_t$ is the weighted sum of all prediction vectors from primary capsules for an object capsule $j$ (line 7 in Pseudocode~\ref{alg:capsule}). 

\begin{equation}
    \label{eq:squash}
     d^{j}_t = \frac{\|v^j_t\|^2}{1+\|v^j_t\|^2}\frac{v^j_t}{\|v^j_t\|}
\end{equation}

\subsection{Pseudocode for Read Attention}

\begin{algorithm}[H]
	\caption{Read Attention} 
	\label{alg:read}
	\begin{algorithmic}[1]
	    \State $h^{dec}_{t-1}$ = previous decoder hidden representation
	    \State $N$= the size of read attention glimpse; $W,H$= (image width, height)
	    \Procedure{ReadAttention}{$x, h^{dec}_{t-1}$}
	    \Comment{Get a N$\times$N read glimpse $g_t$ by applying horizontal and vertical Gaussian filterbank matrices $F_X, F_Y$ to the image}
		    \State Get attention grid parameters: $g_X, g_Y, \log\delta, \log\sigma^2 \leftarrow W_{read\_attention}(h^{dec}_{t-1})$ 
		    \State Scale attention grid centers: $g_X \leftarrow \frac{A+1}{2}(g_X+1)$, $g_Y \leftarrow \frac{A+1}{2}(g_Y+1)$
		    \State Scale attention grid stride:  $\delta \leftarrow \frac{max(A,B)-1}{N-1}\delta$
		    \State Get x, y locations of each grid point $i,j$:
		    \State \hspace{\algorithmicindent} $\mu^i_X = g_X+ (i-N/2-0.5)\delta$
		    \State \hspace{\algorithmicindent} $\mu^i_Y = g_Y+ (j-N/2-0.5)\delta$ 
		    \State Compute Gaussian filterbank matrices ($Z_X$ and $Z_Y$ are normalization constants):
		    \State \hspace{\algorithmicindent} $F_X[i,w] = \frac{1}{Z_X}\exp({-\frac{(w-\mu^i_X)^2}{2\sigma^2}})$
		    \State \hspace{\algorithmicindent} $F_Y[j,h] = \frac{1}{Z_Y}\exp({-\frac{(h-\mu^j_X)^2}{2\sigma^2}})$ 
		    \State Get a read glimpse $g_t$: $g_t \leftarrow F_Y x F^T_X$
		\EndProcedure
	\end{algorithmic} 
\end{algorithm}

\subsection{Pseudocode for Write Attention}
\begin{algorithm}[H]
	\caption{Write Attention} 
	\label{alg:write}
	\begin{algorithmic}[1]
	    \State $h^{dec}_{t}$ = current decoder hidden representation
	    \State $M$= the size of write attention glimpse, $W,H$= (image width, height)
	    \Procedure{WriteAttention}{$h^{dec}_{t}$}
	    \Comment{Reconstruct a M$\times$M write glimpse $w_t$ to the original image size by applying transposed Gaussian filterbank matrices $\hat{F_X}, \hat{F_Y}$}
	        \State Get a write glimpse: $w_t \leftarrow W_{write}(h^{dec}_{t})$
		    \State Get attention grid parameters: $ \hat{g_X}, \hat{g_Y}, \log\hat{\delta}, \log\hat{\sigma}^2 \leftarrow W_{write\_attention}(h^{dec}_{t})$ 
		    \State Scale attention grid centers: $\hat{g_X} \leftarrow \frac{A+1}{2}(g_X+1)$, $\hat{g_Y} \leftarrow \frac{A+1}{2}(g_Y+1)$
		    \State Scale attention grid stride:  $\hat{\delta} \leftarrow \frac{max(A,B)-1}{M-1}\hat{\delta}$
		    \State Get x, y locations of each grid point $i,j$:
		    \State \hspace{\algorithmicindent} $\hat{\mu}^i_X = \hat{g_X}+ (i-M/2-0.5)\hat{\delta}$
		    \State \hspace{\algorithmicindent} $\hat{\mu}^i_Y = \hat{g_Y}+ (j-M/2-0.5)\hat{\delta}$ 
		    \State Compute Gaussian filterbank matrices ($\hat{Z_X}$ and $\hat{Z_Y}$ are normalization constants):
		    \State \hspace{\algorithmicindent} $\hat{F_X}[i,w] = \frac{1}{\hat{Z_X}}\exp({-\frac{(w-\hat{\mu}^i_X)^2}{2\hat{\sigma}^2}})$
		    \State \hspace{\algorithmicindent} $\hat{F_Y}[j,h] = \frac{1}{\hat{Z_Y}}\exp({-\frac{(h-\hat{\mu}^j_X)^2}{2\hat{\sigma}^2}})$ 
		    \State Convert the write glimpse into the original image size: $w_t \leftarrow \hat{F}^T_Y w_t \hat{F}_X$
		\EndProcedure
	\end{algorithmic} 
\end{algorithm}

\newpage
\subsection{Visual Illustrations of Read and Write Attention}
\label{sec:glimpse_attn}
A glimpse, $g_t$, is sampled through application of a grid of N $\times$ N (18$\times$18 in our experiments) Gaussian filters on the input image $x$. The Gaussian filters are generated using four parameters: $g_X, g_Y, \delta, \sigma^2$, which specify the center coordinates of the attention window, the distance between equally spaced Gaussian filters in the grid, and the variance of the filters, respectively. The decoder output $h^{dec}_t$ is linearly transformed into an M $\times$ M write patch $w_t$ (set to 18 $\times$ 18 in our experiments), which is then multiplied by the Gaussian filters to reconstruct the written parts in original image size. 
The Gaussian filters used for the \textit{write} operation differed from those used for the \textit{read} operation.

\begin{figure}[!h]
    \centering
    \includegraphics[width=\textwidth]{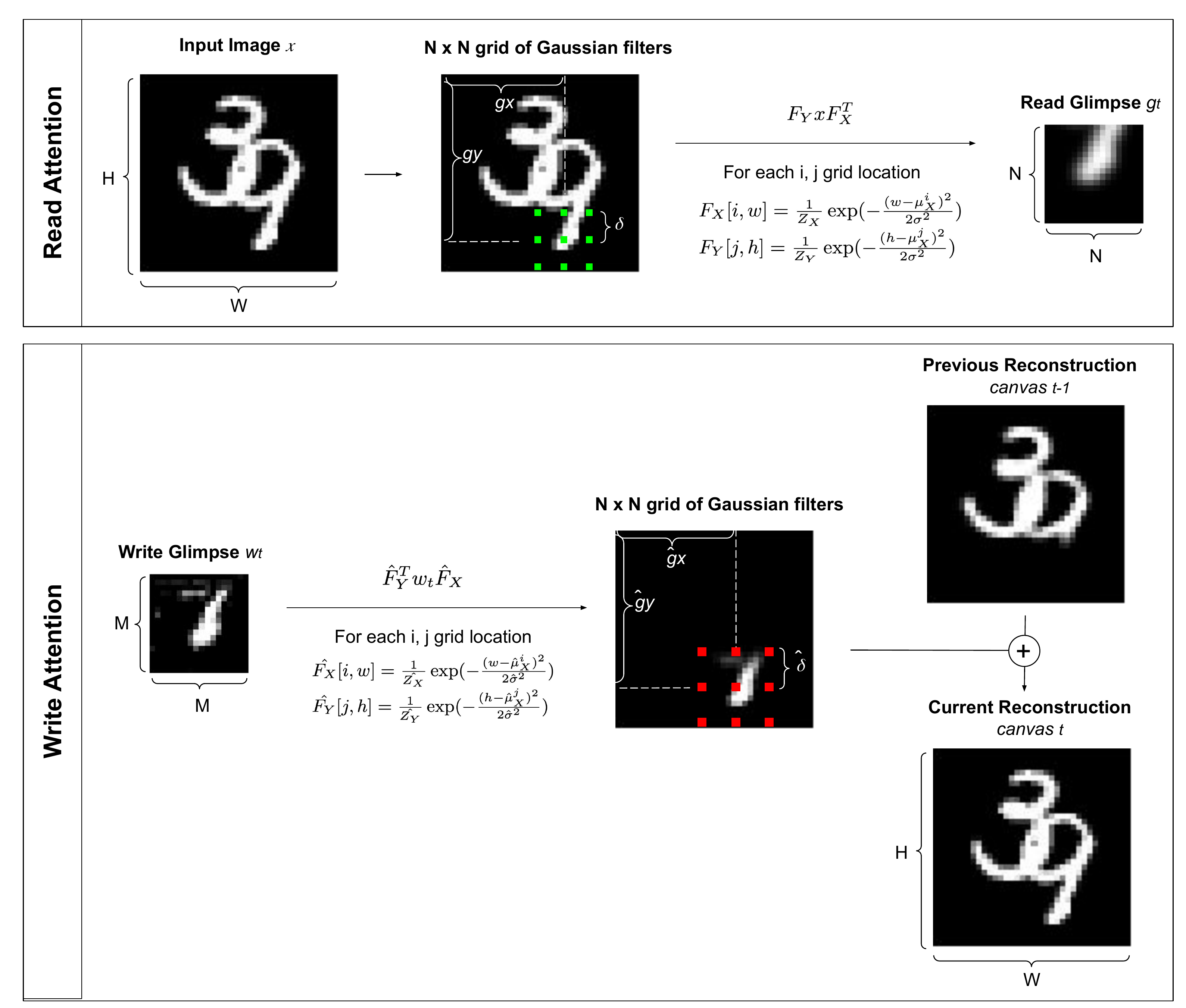}
     \caption{Visual illustrations of read and write attention mechanisms}
    \label{fig:visual_attention}
\end{figure}

\newpage
\subsection{Loss functions}

OCRA outputs object classification scores (cumulative capsule lengths) and image reconstructions (cumulative write canvas). Losses are computed for each output and combined with a weighting $\lambda_{recon}$ (Eq.~\ref{eq:totalloss}). For reconstruction loss, we simply computed the mean squared differences in pixel intensities between the input image and the model's reconstruction. For classification, we used margin loss (Eq.~\ref{eq:marginloss}) so as to give us the flexibility to define ground truth classification that can have multiple objects in each category. This loss ensures that the network yields classification scores similar to ground truth for each image. The loss term has two terms. For each class j, the first term is multiplied by the ground truth $T_{j}$ so this term only would come into play when an object class is present in the ground truth ($L_{present}$ in Pytorch code). Minimizing this loss eventually pushes the model scores for this capsule to be bigger than $(T_{j} -m)$, and with $m$ set at 0.1 this means bigger than .9 when one object is present or bigger than 1.9 when two objects from this category are present in the ground truth. 
The second term is multiplied by a weighting first and then by $\max(0, 1-T_{j})$, which makes this loss come into play only when the class is not present in the ground truth ($L_{absent}$ in Pytorch code). In that case the loss pushes the model scores for this class to be smaller than $m$ ($< 0.1$). These two terms together, when summed across all classes, provide the margin loss for classification. Pytorch implementation of the loss functions are provided below.


\label{sec:loss}

\begin{gather}
    \label{eq:totalloss_supp}
    Total \, Loss =  \sum_{j \in class} Class\,Loss_j + \lambda_{recon} Recon\, Loss \\
    \label{eq:marginloss_supp} 
    \begin{aligned}
    Class \,Loss = \sum_{j \in class} max(0,min(T_j,1)) \cdot max(0, (T_j-m)-\|d_j\|)^2 + \\ \lambda_{absent}\cdot max(0, 1-T_j)\cdot max(0, \|d_j\|-m)^2 
    \end{aligned}
\end{gather}

\begin{lstlisting}[language=Python, caption=Pytorch Implementation of Margin Loss and Reconstruction Loss]
# Margin classification loss 
m = 0.1 # margin of error
lam_abs = 0.5 # down-weighting loss for absent digits

L_present = torch.clamp(y_true, min=0., max=1.) * torch.clamp((y_true-m) - y_pred, min=0.) ** 2 
L_absent = lam_abs * torch.clamp(1 - y_true, min=0.) * torch.clamp(y_pred-m, min=0.) ** 2
L_margin = (L_present+L_absent).sum(dim=1).mean()

# Reconstruction loss 
L_recon = nn.MSELoss()(c, x)  

# Total loss
Loss = L_margin+ lam_recon*L_recon
\end{lstlisting}

\subsection{Measuring accuracy}
When we convert the model scores to multi-object classification, we take into account the thresholds that are set in the loss function. If the model prediction for one class capsule is larger than 1.8 (conservatively selected to be slightly lower than 1.9), this signals the presence of two objects from this class in the image. If no class score is larger than this threshold, the top two highest scores are selected as the model predictions for the two objects in the image.


\clearpage
\subsection{Training Regime and Hyperparameter selection}
\label{sec:training_supp}

OCRA is implemented in Pytorch and the code is available in this \href{https://github.com/Hosseinadeli/OCRA}{repository}. All the model training took place on a single GPU workstation. All the Models trained on the MultiMNIST task were early stopped after 50 epochs of training (taking about 40 hours for the 10glimpse model). The models trained on the MultiMNIST cluttered task were early stopped after 1000 epochs (taking about 40 hours for the 5glimpse model). The models trained on the MultiSVHN task were also stopped after 1000 epochs (taking about 50 hours). We used the Adam optimizer~\citep{kingma2014adam} for all the experiments.

    


\begin{table}[!h]
    \caption{Hyperparameter Setting for MultiMNIST and MultiMNIST Cluttered tasks}
    \label{tab:hypterparam}
    
    \centering    
    \resizebox{\textwidth}{!}{
    \begin{tabular}{lccc}
    \toprule
    Hyperparameters & MultiMNIST-(10/3)glimpse & Cluttered-(5/7)glimpse & MultiSVHN \\ \midrule
    \# timesteps, t & 10/3 & 5/7 & 12\\ 
    \# epoch & 50 & 1000 & 1000\\
    lr &  0.001 & 0.001 & 0.001\\ 
    batch size  & 128 & 128 & 128\\ 
    read glimpse size, $N$ & 18 & 18 & 18\\ 
    write glimpse size, $M$ & 18 & 18 & 18 \\ 
    \# conv1 filters & 32 & 32 & 64 \\ 
    \# conv2 filters & 32 & 32 & 64 \\ 
    lstm size, $dim(h_{enc}), dim(h_{dec})$ & 512 & 512 & 512\\ 
    \# primary capsule &  40 & 40 & 40\\ 
    primary capsule dimension, $dim(p)$ & 8 & 8 & 8\\ 
    \# routings r & 3 & 3 & 3\\ 
    object capsule dimension, $dim(d)$ & 16 & 16 & 16\\ 
    \# background capsules & 0 & 1 & 0\\ 
    reconstruction loss weight, $\lambda_{recon}$ & 10/3 & 175/200 & 200\\ 
    clipping final canvas to [0,1] & TRUE & FALSE & TRUE\\ 
    use reconstruction mask & FALSE & TRUE & FALSE\\ \bottomrule
    \end{tabular}}
\end{table}

Table~\ref{tab:hypterparam} provides all the hyperparameters used for the three tasks. The hyperparameters were mostly the same between the tasks with the few differences. For the MultiMNIST Cluttered task, we added one background capsule and utilized a reconstruction mask, both to allow the model to focus on the main objects and ignore the background clutter. The image is larger (100$\times$100 vs 36$\times$36) in the MultiMNIST Cluttered task, with most of it being empty, the reconstruction loss therefore has a much smaller range compared to the other task. For this reason, and the use of the reconstruction mask, we use a much larger weight for the reconstruction loss to make it comparable to the margin loss. 

For the MultiMNIST task, we clip the cumulative reconstruction canvas to be between [0,1] before comparing it to the input image. This allows the model to overlap different segments in the multi-step process of writing to the canvas without increasing the loss, improving model reconstruction given the high degree of overlap between the digits.

For the MultiSVHN task, we increased the number of the filters in each of  the two convolutional layers to 64.

\clearpage
\section{Additional Results}
\label{sec:results_supp}


\begin{figure}[!h]
    \centering
    \includegraphics[width=\textwidth]{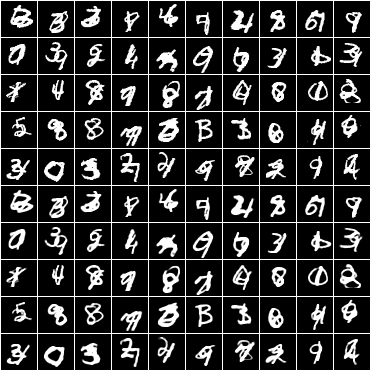}
    \caption{OCRA MultiMNIST output with 10 glimpses, the top 5 rows are the inputs and the bottom 5 rows are model reconstructions.}
    \label{fig:supp_mm_10}
\end{figure}

\begin{figure}[!h]
    \centering
    \includegraphics[width=\textwidth]{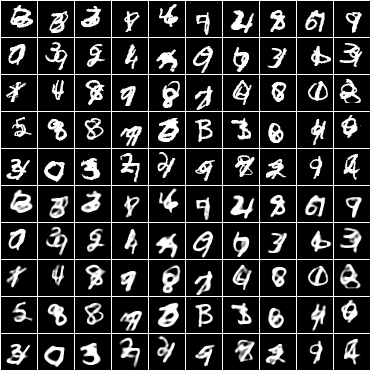}
    \caption{OCRA MultiMNIST output with 3 glimpses, the top 5 rows are the inputs and the bottom 5 rows are model reconstructions.}
    \label{fig:supp_mm_3}
\end{figure}

\begin{figure}[!h]
    \centering
    \includegraphics[width=\textwidth]{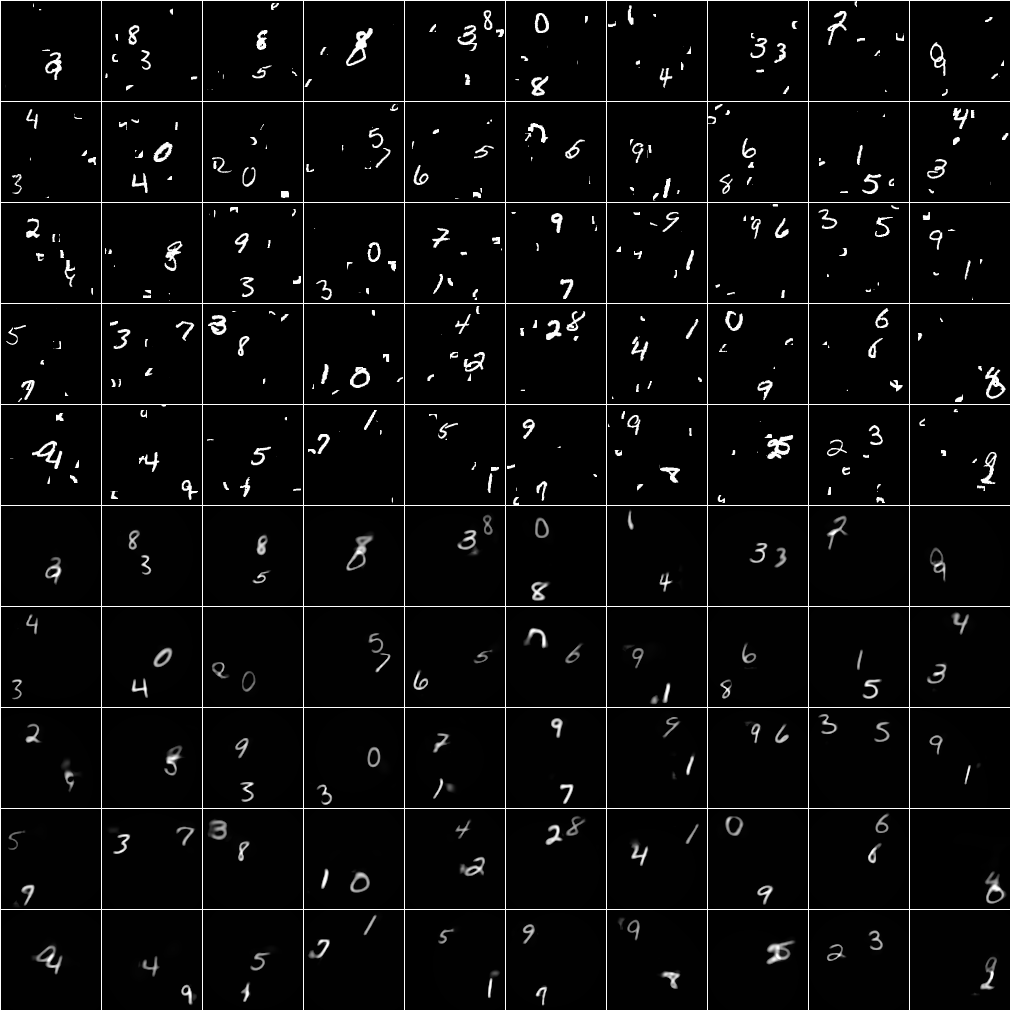}
    \caption{OCRA MultiMNIST Cluttered output with 5 glimpses, the top 5 rows are the inputs and the bottom 5 rows are model reconstructions.}
    \label{fig:supp_mmc_3}
\end{figure}

\begin{figure}[!h]
    \centering
    \includegraphics[height=\textwidth]{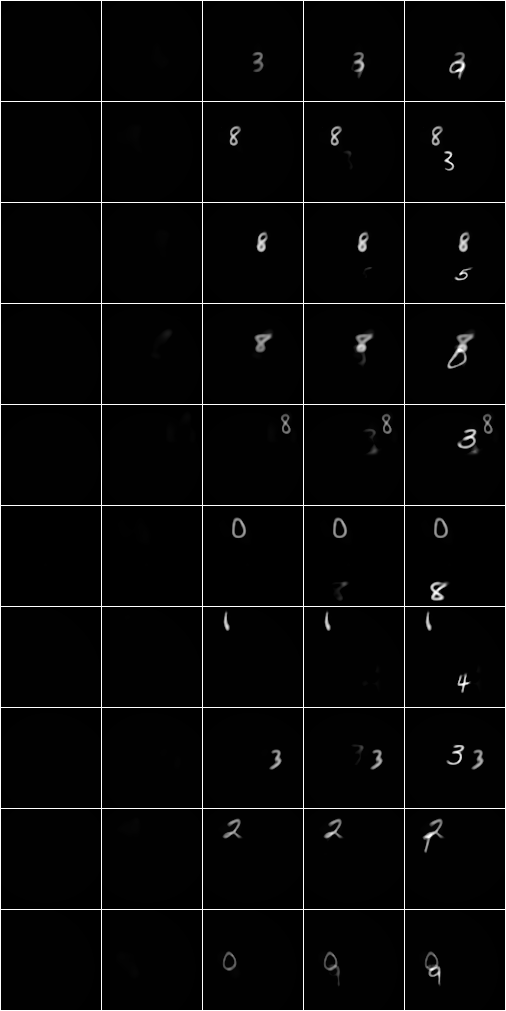}
    \caption{OCRA MultiMNIST Cluttered output with 5 glimpses showing the gradual object-based reconstruction on the cumulative canvas.}
    \label{fig:supp_mmc5steps}
\end{figure}

\begin{figure}[!h]
    \centering
    \includegraphics[width=\textwidth]{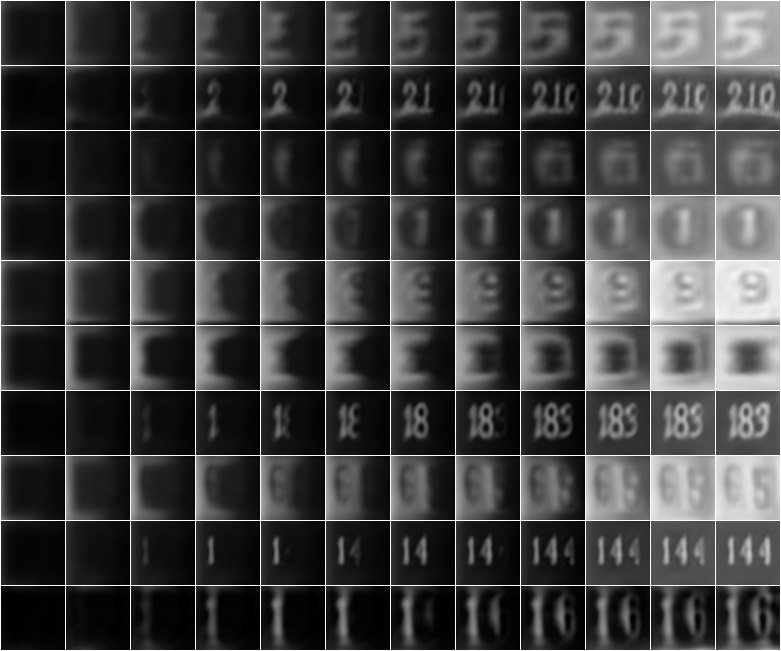}
    \caption{OCRA MultiSVHN output showing the stepwise reconstruction.}
    \label{fig:supp_svhn}
\end{figure}






\clearpage

\end{document}